\documentclass[letterpaper]{article} % DO NOT CHANGE THIS
\usepackage{aaai2026}  % DO NOT CHANGE THIS
\usepackage{times}  % DO NOT CHANGE THIS
\usepackage{helvet}  % DO NOT CHANGE THIS
\usepackage{courier}  % DO NOT CHANGE THIS
\usepackage[hyphens]{url}  % DO NOT CHANGE THIS
\usepackage{graphicx} % DO NOT CHANGE THIS
\usepackage{natbib}  % DO NOT CHANGE THIS AND DO NOT ADD ANY OPTIONS TO IT
\usepackage{caption} % DO NOT CHANGE THIS AND DO NOT ADD ANY OPTIONS TO IT
\usepackage{algorithm}
\usepackage{algorithmic}
\usepackage{array}
\usepackage{longtable}
\usepackage{booktabs}
\usepackage{url}   
\usepackage{pifont}
\usepackage{makecell}
\newcommand{\cmark}{\textcolor{green}{\ding{51}}}
\newcommand{\xmark}{\textcolor{red}{\ding{55}}}
\usepackage{multirow}
\usepackage{cuted}
\usepackage[most]{tcolorbox}
\usepackage{newfloat}
\usepackage{listings}
\DeclareCaptionStyle{ruled}{labelfont=normalfont,labelsep=colon,strut=off} % DO NOT CHANGE THIS
\floatstyle{ruled}
\newfloat{listing}{tb}{lst}{}
\floatname{listing}{Listing}
\title{RealWebAssist: A Benchmark for Long-Horizon Web Assistance with Real-World Users}
\author{
    Suyu Ye\equalcontrib \textsuperscript{1},
    Haojun Shi\equalcontrib \textsuperscript{1},
    Darren Shih \textsuperscript{1},
    Hyokun Yun \textsuperscript{2},
    Tanya G. Roosta \textsuperscript{2},
    Tianmin Shu \textsuperscript{1}
}
\affiliations {
    \textsuperscript{\rm 1}Johns Hopkins University, \\
    \textsuperscript{\rm 2}Amazon.com\\
    \{sye10, hshi33, dshih5, tianmin.shu\}@jhu.edu, \{yunhyoku,troosta\}@amazon.com
}

\begin{document}

\maketitle

\begin{abstract}
To achieve successful assistance with long-horizon web-based tasks, AI agents must be able to sequentially follow real-world user instructions over a long period. Unlike existing web-based agent benchmarks, sequential instruction following in the real world poses significant challenges beyond performing a single, clearly defined task. For instance, real-world human instructions can be ambiguous, require different levels of AI assistance, and may evolve over time, reflecting changes in the user's mental state. To address this gap, we introduce RealWebAssist, a novel benchmark designed to evaluate sequential instruction-following in realistic scenarios involving long-horizon interactions with the web, visual GUI grounding, and understanding ambiguous real-world user instructions. RealWebAssist includes a dataset of sequential instructions collected from real-world human users. Each user instructs a web-based assistant to perform a series of tasks on multiple websites. A successful agent must reason about the true intent behind each instruction, keep track of the mental state of the user, understand user-specific routines, and ground the intended tasks to actions on the correct GUI elements. Our experimental results show that state-of-the-art models struggle to understand and ground user instructions, posing critical challenges in following real-world user instructions for long-horizon web assistance.
\end{abstract}

\section{Introduction}

As an integral part of people's daily life, many of our everyday tasks are performed on the internet. With the tremendous advances in open-ended agents driven by large reasoning models (LRMs) and vision-language models (VLMs), there has been increasing interest in engineering web-based agents that can assist humans with complex tasks on the web following humans' instructions \citep{zheng2024gpt, nakano2022webgptbrowserassistedquestionansweringhuman}. Recent works have demonstrated the promising performance of web-based agents on planning \citep{putta2024agent, wang2024agent, yao2023reactsynergizingreasoningacting} and Graphical User Interface (GUI) grounding \citep{cheng2024seeclick, wu2024atlas, gou2024navigating, yang2024aria, xu2024aguvisunifiedpurevision}, across diverse websites, tasks, and GUI interfaces. 

Despite these encouraging results, there have not been systematic studies on long-horizon web assistance with real-world users. Existing benchmarks (e.g., \cite{zhou2023webarena, deng2024mind2web, cheng2024seeclick, yao2022webshop, jang2024videowebarena}) typically focus on performing a task based on a single instruction. Additionally, the instructions in the current benchmarks were not collected from real users during natural web use sessions, lacking the realism of real user instructions. As a result, these benchmarks do not capture the full complexity of real users' web behavior and instructions. 

\begin{figure*}[t!]
    \centering
    \includegraphics[trim=0cm 17cm 0cm 0cm, clip, width=0.95\textwidth]{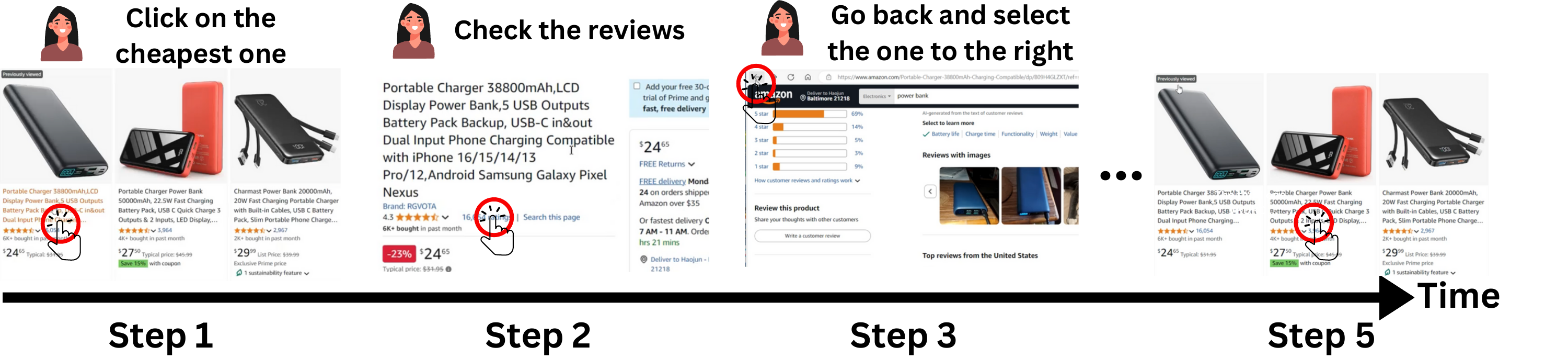}
    \caption{An example sequential instruction following task with a real-world user. The red circles indicate the correct actions based on the user's spoken instructions. Sequential instructions introduce unique challenges, such as the need to retain and reason over past context. For instance, the instruction in step 3 requires information from step 1 to be correctly interpreted.}
    \label{fig:sequential_instruction}
\end{figure*}

\begin{figure*}[t!]
    \centering
    \includegraphics[trim=0cm 18.7cm 10.5cm 0cm, clip, width=0.9\textwidth]{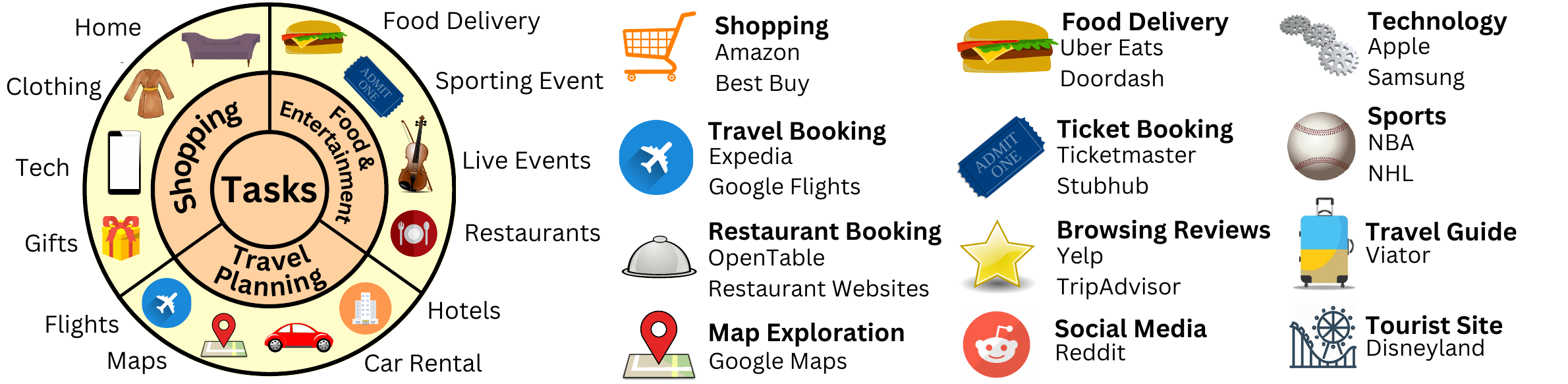}
    \caption{Examples of general task categories (left) and websites visited (right) in RealWebAssist. The tasks span a wide range of real-world scenarios, from shopping to food \& entertainment to travel planning, which encourages users to visit many different websites.}
    \label{fig:websites_and_tasks}
\end{figure*}

To bridge this gap, we propose \textbf{RealWebAssist}, the first sequential instruction following benchmark that evaluates long-horizon web assistance with real-world users. As illustrated in Figure~\ref{fig:sequential_instruction}, to perform a task, a user will instruct an AI assistant in a long sequence. Based on the past instructions and screenshots, the AI assistant must execute one or a few steps of actions to perform the latest instruction. Additionally, a user can engage in repeated interactions over a series of tasks with the assistant in a long session up to 40 minutes. To construct RealWebAssist, we recruited real users to instruct an assistant to perform multiple real-world tasks on the web. We created a large dataset with real user instructions (in both speech and text) for diverse real-world tasks and websites (as shown in Figure~\ref{fig:websites_and_tasks}).

The sequential instruction following tasks in our RealWebAssist benchmark reflect the natural human behavior on the web. First, real-world users may not initially know what they are looking for. Thus, they need to engage in information seeking on multiple web pages (e.g., step 1-2 in Figure~\ref{fig:sequential_instruction}), sometimes even across websites. Second, based on new information such as product reviews, users may change their minds (e.g., step 3). Third, users give simple instructions that are seemingly ambiguous out of the context but could be interpreted based on spatial and temporal context via pragmatic reasoning \citep{goodman2016pragmatic, fried2023pragmaticslanguagegroundingphenomena}. For instance, the third instruction in Figure~\ref{fig:sequential_instruction} does not explicitly describe which product, but an intelligent assistant should be able to infer the true user intent and correctly select the product in the user's mind. Lastly, in our benchmark, users can browse the websites and have the autonomy to make critical decisions (such as purchasing) on their own, which is complementary to existing benchmarks that focus on agents' planning ability to fully complete the tasks without human involvement.

We systematically evaluate state-of-the-art models, including GUI grounding, VLMs, and large reasoning models. Experimental results reveal that these models lack several key abilities, including grounding, understanding user intents, reasoning about spatial and temporal context, and adapting to user-specific routines.

\section{Related Works}

\begin{table*}[htbp!]
  \begin{center}
    \begin{scriptsize}
    \renewcommand{\arraystretch}{1}  % 
    \setlength{\tabcolsep}{4pt}       % Adjust column spacing for compactness
    \begin{tabular}{p{4.2cm} >{\centering\arraybackslash}p{1.3cm} >{\centering\arraybackslash}p{1.3cm} >{\centering\arraybackslash}p{1.3cm} >{\centering\arraybackslash}p{1.3cm} >{\centering\arraybackslash}p{1.3cm} >
    {\centering\arraybackslash}p{1.3cm}}
    \toprule
    \textbf{Benchmark} & \textbf{Real User} & \textbf{Sequential\newline Instructions} & \textbf{Real\newline Websites} & \textbf{GUI\newline Grounding} & \textbf{Speech} & \textbf{\# Instructions}\\
    \midrule
    SreenSpot \citep{cheng2024seeclick} & \xmark & \xmark & \cmark &\cmark & \xmark & 1200+ \\
    WebArena \citep{zhou2023webarena} & \xmark & \xmark & \xmark & \xmark & \xmark & 812\\
    Mind2Web \citep{deng2024mind2web} & \xmark & \xmark & \cmark & \xmark & \xmark & 2000+\\
    %MoTIF \citep{burns2022dataset} & \cmark & \xmark & \xmark & \xmark & 4707\\
    WebLINX \citep{lu2024weblinx} & \xmark & \cmark & \cmark & \xmark & \xmark & 512
    \\
    VideoWebArena \citep{jang2024videowebarena} & \xmark & \xmark & \xmark & \xmark & \cmark & 2021
    \\
    WebShop \citep{yao2022webshop} & \xmark & \xmark & \xmark & \xmark & \xmark & 12087 
    \\
    BearCubs \citep{song2025bearcubs} & \xmark & \xmark & \cmark & \xmark & \xmark & 111
    \\
    \midrule
    \textbf{RealWebAssist (Ours)} & \cmark & \cmark & \cmark & \cmark & \cmark & 1885 \\
    \bottomrule
    \end{tabular}
    \end{scriptsize}
  \end{center}
\caption{Comparison between RealWebAssist and existing web agent benchmarks on several key aspects: (1) whether instructions were given by real-world users instead of annotators, (2) whether there is a sequence of instructions, (3) whether there are real-world websites, (4) whether the agent needs to execute actions by selecting coordinates on webpages, (5) whether the instructions are speech instructions, and (6) the number of total instructions.}
  \label{tab:web_benchmarks}
\end{table*}

\textbf{Web Agent Benchmarks.}
Existing web agent benchmarks primarily evaluate the performance of web agents on tasks with clearly defined, unambiguous instructions, often overlooking the complexities of real-world users' behavior and their instructions to an AI assistant. On WebArena \citep{zhou2023webarena}, Mind2Web \citep{deng2024mind2web}, and WebShop \citep{yao2022webshop}, an agent follows a single instruction to perform an isolated task. While they offer an evaluation of an agent's planning capacity, they lack the evaluation of an agent's ability to follow a long sequence of user instructions on long-horizon web tasks. There have also been GUI grounding benchmarks, such as ScreenSpot \citep{cheng2024seeclick}, that focused on grounding simple instructions to clicking actions on webpages. These instructions only instruct web agents to click web elements rather than reaching a user goal (e.g., purchasing an item).  WebLINX \citep{lu2024weblinx} features sequential instruction following. However, the instructions were generated by annotators who received detailed guidelines and extensive training, rather than by actual users. The resulting instructions do not capture the nuances and complexity of real-world user instructions that naturally emerge in interactions with an assistant. In contrast, RealWebAssist consists of sequential instruction following tasks for assisting real-world users, providing a novel set of challenges necessary for long-horizon web assistance for real-world users. Table~\ref{tab:web_benchmarks} summarizes key differences between RealWebAssist and prior benchmarks.

\textbf{Autonomous Web Agents.}
There have been many recent works on engineering autonomous web agents through retrieval augmented planning \citep{kim2024rada, zhou2024languageagenttreesearch,wu2024oscopilotgeneralistcomputeragents, he2024webvoyagerbuildingendtoendweb, pan2024autonomousevaluationrefinementdigital}, finetuning \citep{hong2024cogagentvisuallanguagemodel, gur2024realworldwebagentplanninglong, deng2024mind2web, pang2024iterativereasoningpreferenceoptimization, zhang2024lookscreensmultimodalchainofaction}, learning workflows \citep{zhang2023appagentmultimodalagentssmartphone, wang2024agent, zheng2024synapsetrajectoryasexemplarpromptingmemory, majumder2023clincontinuallylearninglanguage, cai2024largelanguagemodelstool}, reinforcement learning \citep{liu2018reinforcement, shi2017world, nogueira2016end, humphreys2022data}, and combinations of these methods \citep{liu2023bolaa, putta2024agent}. These works focus on planning for a single task. However, there has not been much work on understanding and following real-world users' sequential instructions on long-horizon tasks.

\textbf{GUI Grounding.} One key ability for web agents in many assistance tasks is to ground instructions to clicking actions on a webpage. Recent works have explored VLM finetuning (e.g., \cite{gou2024navigating, wu2024atlas, yang2024aria, yang2025gta1, wu2025gui, qin2025ui, xu2025aguvisunifiedpurevision, yuan2025enhancingvisualgroundinggui}) as well as prompting pretrained VLMs with segmentations of web elements (e.g., \cite{yang2023setofmark}) for enabling GUI grounding. These methods generate coordinates or bounding boxes on webpages to indicate where to click. They have only been trained on low-level instructions that clearly refer to web elements. It remains unclear if they can understand real-world user instructions that must be interpreted considering context or may refer to high-level goals.

\section{RealWebAssist Benchmark}

\subsection{Problem Setup}
RealWebAssist evaluates agents' ability to follow long-horizon, sequential web instructions to assist users with their high-level goals. In each task, a human user will try to reach an open-ended goal such as ``buy formal outfits for a formal event" by instructing the assistant through a series of spoken instructions. The dataset is collected from interactions between human users and human assistants in a human experiment. To evaluate agents, we use the human assistants' actions to evaluate the agents' success.

\begin{figure}[t!]
    \centering
    \includegraphics[trim=0.2cm 21.8cm 13.2cm 0cm, clip, width=0.9\textwidth]{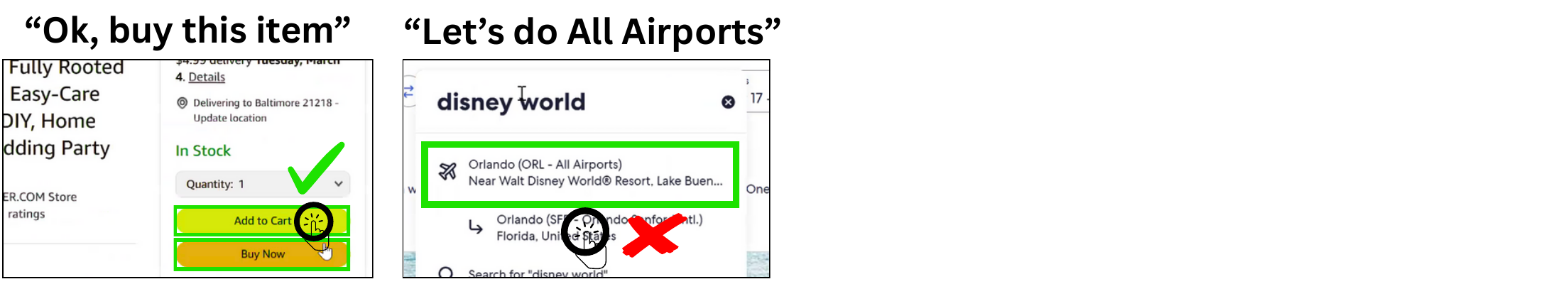}
    \caption{Multiple actions can satisfy a user’s intent. A web agent's action is considered correct if the coordinate they provide is within one of the annotated correct regions.}
    \label{fig:correct_wrong_example}
\end{figure}

In RealWebAssist, a web agent has access to the current instruction, webpage (as a screenshot), and all the past interactions (previous instructions \& screenshots of webpages). Since we are focusing on tasks on real-world websites, it is challenging to ensure safety and reproducibility in an interactive evaluation setting. Therefore, we adopt an offline evaluation setting following prior web-based agent benchmarks with real websites \citep{deng2024mind2web, cheng2024seeclick}. Specifically, for each instruction collected from the human experiment, the agent needs to identify the correct element to interact with by providing a coordinate or a bounding box to click on the webpage. As shown by figure \ref{fig:correct_wrong_example}, a web agent's action is considered correct if the coordinate or the center of the bounding box they provide falls in the annotated correct regions on the webpage. If there are multiple steps corresponding to one instruction, we evaluate if the web agent's actions for the same instruction are all correct.

\subsection{Evaluation Metrics}
We consider the following evaluation metrics:
\begin{itemize}\setlength\itemsep{0pt}
\setlength{\parskip}{0pt}
\setlength{\parsep}{0pt}
    \item \textbf{Task success rate:} A task is successful if the web agent can correctly produce actions for all instructions in a task.
    \item \textbf{Average progress:} We measure the progress of a task by the percentage of consecutive instructions the web agent can successfully perform before its first error in the task.
    \item \textbf{Step success rate:} We also consider a teacher forcing setting as a simpler, diagnostic evaluation, where the web agent will only need to follow the instruction at a single step of a task assuming all previous instructions have been successfully performed. 
\end{itemize}

\subsection{Dataset Construction}

\textbf{Setup.} We recruited 10 participants (4 female, 6 male, mean age = 20 years) from a US university campus, none of whom had prior knowledge of the study's purpose, to construct the dataset. All participants were native or fluent English speakers. Each participant completed a 40-minute real-world web assistance session in which they tackled a series of open-ended tasks designed to encourage diverse strategies.  During each session, participants verbally instructed an experimenter, who operated the computer on their behalf, to complete the tasks. We captured screen recordings and used a high-quality USB microphone to record speech as raw data. The user study was approved by an institutional review board. 

\textbf{User Tasks.} To increase the instruction diversity and realism, participants received general web-based tasks requiring active information seeking, sub-goal planning, and comparison among various options. We generated the task list by few-shot prompting GPT-4o with open-ended tasks, followed by manual filtering and editing to ensure task quality and feasibility. These tasks provide only general guidance, ensuring flexibility for personal decision-making. Example tasks include ``Purchase an outfit for a formal event" and ``Plan a 5-day trip to Japan, booking both flights and hotels". Each user finishes about 10 tasks.

\textbf{Emergent User Behavior.} In our realistic, open-ended settings,  users exhibit rich behaviors that are not present in previous benchmarks. These include, but are not limited to, information seeking, researching and comparing different options, change of mind, and trial-and-error.

\textbf{Annotations.}
We manually labeled RealWebAssist data to ensure high-quality annotations. We first segmented the full recording into individual clips corresponding to each user's instructions. In our benchmark, we disregard user speech unrelated to explicit instructions for the assistant, such as filler words or verbalized thought processes. For each instruction, we provide raw speech, speech transcript, webpage, and the correct regions to click (in the form of one or more bounding boxes). When there were multiple correct answers for the instructions (for instance, ``can you close all the current tabs"), we annotated all correct regions with multiple bounding boxes. When the experimenter made a mistake during the data collection sessions, we annotated the correct action intended by the user. If an instruction required multiple steps to complete, we set the instruction at each step as the same instruction. To generate the text instructions, we used an off-the-shelf recognition model, Whisper Large-V3 \citep{radford2023robust}, to transcribe users' speech and then manually fixed transcription errors. For all the instructions, we have three annotators verifying all of them, ensuring 100\% agreement.

\textbf{Dataset Statistics.} RealWebAssist contains 1,885 user instructions across 107 tasks, 66 websites, and 2,524 screenshots. In addition to the benchmark, we also plan to release the raw data, consisting of over 6 hours of video \& audio. 
 
\subsection{Key Challenges}

\begin{figure*}[t!]
    \centering
    \includegraphics[trim=13cm 2cm 13cm 2cm, clip, width=1.0\textwidth]{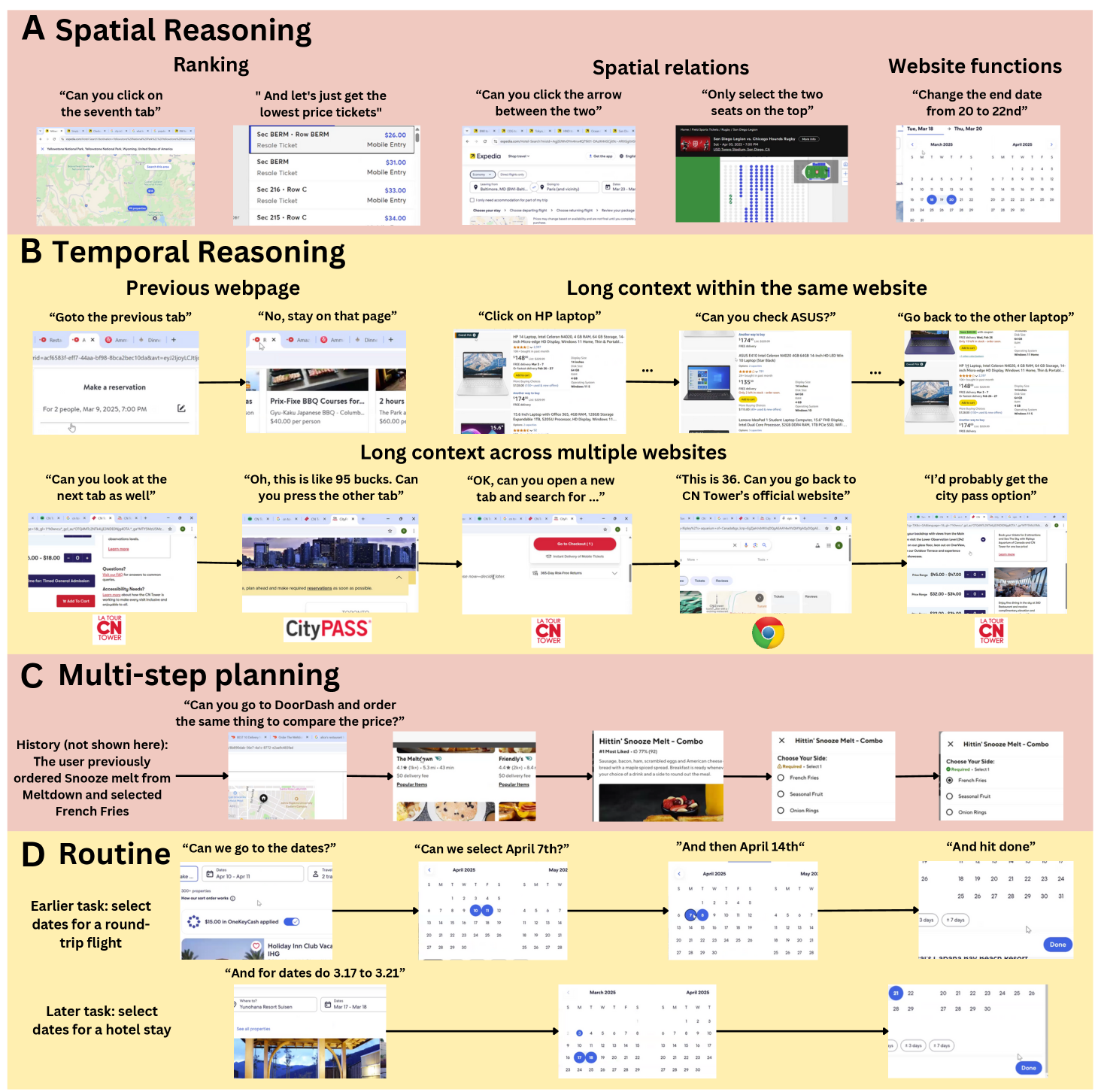}
    \caption{Key challenges introduced by RealWebAssist: (A) spatial reasoning, (B) temporal reasoning, (C) multi-step planning, and (D) learning user-specific routines.}
    \label{fig:Benchmark_figure}
\end{figure*}

RealWebAssist features multiple challenges as illustrated in Figure~\ref{fig:Benchmark_figure}, including spatial and temporal reasoning needed to understand ambiguous and context-dependent user instructions, planning for multiple steps of actions to reach the goal communicated by an instruction, and learning about user-specific routines. These key challenges provide a more realistic and holistic evaluation of a web agent's reasoning, planning, and learning abilities to assist real-world users on long-horizon tasks. It is worth noting that many of these challenges, in particular, spatial reasoning, temporal reasoning, and routine understanding, are not present in existing web agent benchmarks. Unlike RealWebAssist, prior benchmarks, such as ScreenSpot \citep{cheng2024seeclick}, WebArena \citep{zhou2023webarena}, and Mind2Web \citep{deng2024mind2web}, only include clear, unambiguous, and non-sequential instructions.

\textbf{Spatial Reasoning.} When referring to one of the elements on a webpage, real-world users tend to use a concise instruction that can be understood conditioned on spatial context instead of an overly elaborated instruction. For instance, when instructing an assistant to buy a product, users may give short instructions such as ``select the cheapest one,'' instead of describing the desired product in detail. Figure ~\ref{fig:Benchmark_figure}A depicts different types of spatial reasoning that rely on diverse spatial contexts, including ranking, spatial relations, and overall website functionalities. It is worth noting that these instructions may sometimes reveal users' preferences (e.g., preferred seating), providing additional information for the web agent to provide potentially more customized assistance in the future.

\textbf{Temporal Reasoning.} In our sequential instruction following tasks, users may instruct an assistant with the history as an assumed temporal context. For example, to understand the intended meaning of ``click the last item," the assistant must memorize the items the user has viewed in the past. Figure \ref{fig:Benchmark_figure}B shows temporal reasoning based on different kinds of temporal context, ranging from short context between two consecutive webpages to long context with the same website to long context across websites. From the temporal context, the assistant needs to memorize crucial elements in the previous webpages, infer and track a user's mind (e.g., change of mind about what to buy) based on the past instructions and webpages, and identify the earlier webpage the user refers to. Such temporal reasoning has not been evaluated in prior web agent benchmarks. However, it is very common in our benchmark due to the nature of human web browsing behavior as well as human instructions guided by pragmatics \citep{goodman2016pragmatic}. 

\textbf{Multi-step Planning.} Many instructions require multiple steps to complete. In these cases, the assistant needs to interpret the goal implied by the instruction and plan a sequence of actions to achieve that goal. This goes beyond grounding the instruction to a single action on the current webpage. Figure~\ref{fig:Benchmark_figure}C shows an example where the agent was asked to repeat the same order on another food delivery website to check if the price would be different. A successful execution of this instruction would require the agent to first understand what the order is to ground the goal on the current website and generate a successful multi-step plan.

\textbf{Routine.} Since our benchmark allows a user to engage in repeated interactions with an assistant over multiple tasks, we observe that users may define routines understood by the assistant after repeated interactions. As shown in Figure \ref{fig:Benchmark_figure}D, the user initially gave detailed step-by-step instructions when selecting arrival and departure dates for a flight. In a subsequent task, however, the user simplified them into a single instruction when selecting dates for a hotel room. Such shorter instructions become possible after establishing a routine in the earlier task. Cognitive studies found that procedural abstraction, like these routines, naturally emerges in human cooperative communication through repeated interactions, allowing more efficient communication with partners \citep{mccarthy2021learning}. The emergence of such routines in our benchmark poses a novel challenge for web agents—learning user-specific procedural abstraction via repeated interactions to achieve human-like adaptive assistance. We hypothesize that this ability could enhance users' perception of the AI assistant, as it understands human cooperative communication.

\section{Experiments}

\subsection{Baselines}

We evaluated several types of models for web agents commonly evaluated in existing web agent benchmarks that have real-world websites (i.e., offline evaluation). For all the experiments, we use the ground-truth captions for instructions.

\textbf{GUI Grounding Models.} GUI grounding models directly translate an instruction to an action on a webpage. There are two general types of grounding models. First, Set-of-Mark (SoM) \citep{yang2023setofmark} segments salient elements on a webpage using an off-the-shelf segmentation model (e.g., SAM \citep{kirillov2023segment} and Semantic-SAM \citep{li2023semantic}) and prompts a VLM to select a segment mask to identify the clicking area corresponding to the given instruction. Second, VLMs finetuned on datasets with paired instructions and annotated clicking coordinates or bounding boxes. We evaluated UGround-V1 \citep{gou2024navigating}, OS-Atlas \citep{wu2024atlas}, Aria-UI \citep{yang2024aria}, GTA-1 \citep{yang2025gta1}, GUI-Actor \citep{wu2024oscopilotgeneralistcomputeragents}, and UI-TARS \citep{qin2025ui}.

\textbf{VLM/LRM + Grounding.}
Grounding models are designed or trained to ground a simple instruction to a webpage and thus tend to lack reasoning or planning capabilities. To address this, we leveraged VLMs and LRMs to first translate real user instructions to more understandable ones for grounding models. In particular, a VLM or an LRM needs to reason about the true user intent implied by the instruction and the spatial \& temporal context. For instructions that require multiple actions, it needs to generate a plan to complete the instructions. Finally, it needs to generate a straightforward, clear instruction for the grounding model to produce the final action at each step. We evaluated state-of-the-art VLMs \citep{OpenAI2023GPT4TR, comanici2025gemini25pushingfrontier, qwen2025qwen25technicalreport}, as well as state-of-the-art LRMs \citep{jaech2024openai, comanici2025gemini25pushingfrontier, claude37sonnet2024}. In the main results, we paired each VLM and LRM with the grounding model that achieved the highest step accuracy (GTA-1). For all VLMs and LRMs, we provide the past 10 steps for context, which we found to be a reasonable fixed context length in our preliminary study, balancing cost and informativeness. We also found that prompting models with screenshots of past webpages could incur a high cost. Therefore, we only prompt the models with the screenshot of the current webpage. For the history, we prompted GPT-4o to generate text-based action history based on consecutive screenshots and the instructions at each step. We then used this text-based history description for the evaluated VLMs and LRMs.

\textbf{Finetuning.}
To evaluate whether models can learn to better follow real-world user instructions with additional training, we finetuned the best-performing grounding model (GTA-1) following the model's original group relative policy optimization (GRPO) training procedure \citep{yang2025gta1} on 9 participants' data and tested it on the held-out participants' instructions. Specifically, we trained the grounding model to produce an action based on the past 10 steps of actions (in text), the current webpage screenshot, and the instruction. We enumerated different train/test splits and reported the averaged performance, either using the finetuned model alone or pairing it with the best VLM or LRM.

\subsection{Results}

% Main categories table
\begin{table*}[t!]
    \centering
    \begin{tabular}{p{2.5cm}|p{5cm}|>{\centering\arraybackslash}p{1.8cm}|
    >{\centering\arraybackslash}p{1.8cm}|
    >{\centering\arraybackslash}p{1.8cm}}
        \hline
        \textbf{Category} & \textbf{Model} & \makecell{\textbf{Task} \\ \textbf{Success}} & \textbf{Progress} & \makecell{\textbf{Step} \\ \textbf{Accuracy}} \\  
        \hline
        \multirow{1}{2.5cm}{Human}
                  & Human Operator & 93.4 & 96.4 & 99.2 \\
        \hline
        \multirow{4}{2.5cm}{Grounding}
                  & Set-of-Mark & 0.0 & 2.7 & 29.8 \\
                  & OS-Atlas & 0.0 & 3.8 & 26.6 \\
                  & Aria-UI & 0.0 & 2.4 & 32.8 \\
                  & UGround-V1 & 0.0 & 6.2 & 47.7 \\
                  & UI-TARS & 2.8 & 13.1 & 53.8 \\
                  & GTA-1 & 3.7 & 17.7 & 61.5 \\
                  & GUI-Actor & 5.7 & 14.7 & 61.4 \\
        \hline
        \multirow{3}{2.5cm}{VLM + Grounding}
                & GPT-4o + GTA-1 & 8.4 & 23.5 & 72.7 \\
                & Qwen 2.5 72B + GTA-1 & 9.3 & 24.3 & 69.0 \\
                & Gemini 2.5 Flash + GTA-1 & 11.2 & 26.9 & 75.4 \\
        \hline
        \multirow{3}{2.5cm}{LRM + Grounding}      
                  & o1 + GTA-1 & 7.5 & 17.7 & 68.2 \\
                  & Gemini 2.5 Pro + GTA-1 & 8.4 & 23.5 & 74.5 \\
                  & o4-mini + GTA-1 & 10.3 & 21.7 & 67.1 \\
                  & Claude 3.7 Sonnet + GTA-1 & 12.1 & 26.7 & 68.8 \\
                  & o3 + GTA-1 & \textbf{14.0} & \textbf{28.7} & \textbf{76.7} \\
        \hline\hline
        \multirow{3}{2.5cm}{Finetuned}
                  & GTA-1-F & 3.7 (+0.0) & 19.7 (+2.0) & 64.3 (+2.8) \\
                  & Gemini 2.5 Flash + GTA-1-F & 11.2 (+0.0) & 26.9 (+0.0) & 75.4 (+0.0)\\
                  & o3 + GTA-1-F & \textbf{14.0} (+0.0) & \textbf{28.7} (+0.0) & \textbf{76.7} (+0.0) \\
        \hline
    \end{tabular}
    \caption{Model Performance including task success rate, average progress, and step accuracy. All results are in \%. The best performance of pretrained models and finetuned models is highlighted in bold. GTA-1-F indicates the finetuned GTA-1. Plus sign indicates the improvement compared to using the raw model for the same set of instructions.}\label{tab:categories}
\end{table*}

Main results are summarized in Table \ref{tab:categories}. All models fell short in following real user instructions. The highest task success rate was only 14.0\%, and the highest average progress was only 28.7\%, a large gap compared to humans (93.4\% task success rate). This difference has a 95\% confidence interval of [71.3, 87.5], and is highly significant with p-value $<$ 0.0001. Grounding methods by themselves failed to finish most tasks. However, when paired with the best-performing grounding model (GTA-1), instructions generated by VLMs \& LRMs significantly improved the performance. LRMs performed marginally better than most VLMs. Across all three metrics, Gemini 2.5 Flash, Gemini 2.5 Pro, and o3 showed the strongest performance. Finetuning GTA-1 on real user data marginally improved its performance, but finetuning offered no benefit when GTA-1 was paired with VLMs and LRMs, since the finetuned model is trained to adapt to real users' instructions instead of instructions generated by VLM or LRM. 

%\newpage
%\begin{figure}[h]
  %\centering \includegraphics[width=0.4\textwidth]{figures/context_length_accuracy.png}
  %\caption{Effect of context length on GPT-4o + UGround-V1. Dotted lines represent %results with full context.}
  %\label{fig:context_length}
%\end{figure}

%Additionally, we evaluated the best-performing VLM (GPT-4o) + UGround-V1 with varying history context lengths, from no history to full interaction history with the same user, which can be up to 305 steps. An ideal assistant should be able to leverage different kinds of historical context based on different instructions, ranging from no history to multi-task history context (e.g., for routine learning). As shown in Figure \ref{fig:context_length}, increasing context length also does not necessarily lead to better performance. GPT-4o + UGround-V1 achieved the highest task success rate with a context length of 10, and increasing the context length further led to poorer performance. Additionally, providing all past instructions and actions of a user as context is not only expensive, since the context may have hundreds of steps, but also does not increase performance, indicating that the context is not being effectively used. It also suggests a lack of effective routine learning ability. All baseline experiments used the ground truth transcripts of user speech instructions to avoid introducing errors caused by speech recognition. %We provide the results with the speech recognition results as input in Appendix~\ref{appendix:speech_to_text}.

\section{Discussion}

\begin{figure*}[t!]
    \centering
    \includegraphics[trim=2.5cm 8cm 2.5cm 3cm, clip, width=1.0\textwidth]{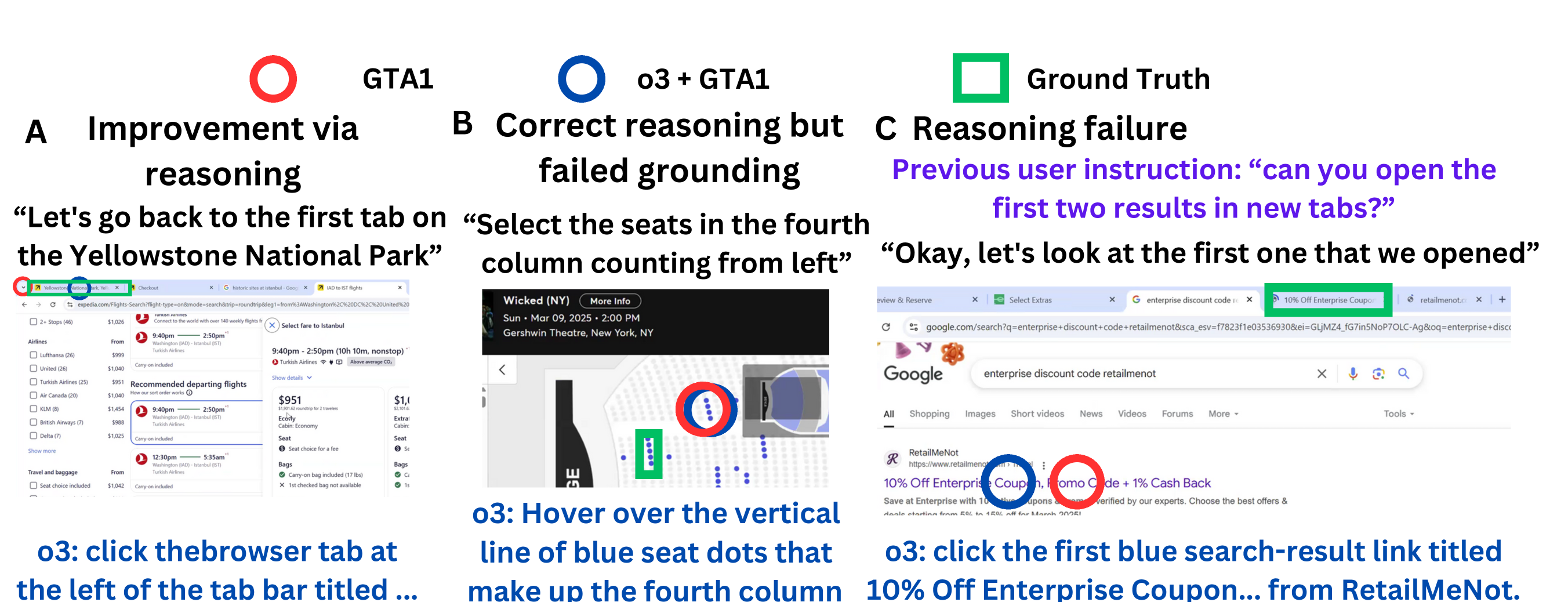}
    \caption{Qualitative results. The captions show instructions generated by o3 (the best LRM). (A) Error corrected by using o3 to convert instructions. (B) Failure caused by GTA-1 when o3 reasons correctly. (C) Reasoning failure caused by o3.}
    \label{fig:Failures_figure}
\end{figure*}

\textbf{Can grounding models understand real-world user instructions?}
There remains a significant gap in the performance of current direct grounding methods. The best grounding model, GUI-Actor, has a task success rate of only 5.7\%. Figure \ref{fig:Failures_figure} illustrates various failure cases encountered when directly using GTA-1. Unsurprisingly, grounding models fail to interpret instructions requiring reasoning due to their limited reasoning capabilities. However, even for context-free instructions involving straightforward spatial reasoning—tasks where grounding methods should excel—they frequently misinterpret spatial layouts or rankings. For instance, they often incorrectly select elements for instructions such as ``click the first one.''

\textbf{How can VLMs \& LRMs help?}
VLMs or LRMs can convert the original user instructions into more direct and explicit descriptions that a grounding model can more easily understand. This is made possible by their reasoning capacities. For instance, in Figure \ref{fig:Failures_figure}A, the grounding model (GTA-1) on its own fails to select the first tab: it selects the first element instead of the first tab. However, it succeeds after o3 rewrites the instruction to refer to the title. As shown in Figure \ref{fig:Failures_figure}B, grounding models may sometimes still fail due to inherent limitations even when VLMs/LRMs generate clearer instructions. Nonetheless, incorporating VLMs or LRMs significantly improves overall performance.

\textbf{What are the limitations of VLMs \& LRMs?}
While VLMs and LRMs help, the highest task success rate is still only 14.0\%. Beyond errors from grounding models (e.g., Figure~\ref{fig:Failures_figure}B), they continue to struggle with complex temporal reasoning. In Figure~\ref{fig:Failures_figure}C, the user previously asked to open the first two search results in new tabs. When later instructed to “look at the first one we just opened,” o3 failed to identify which element “the first one” referred to—instead of the first newly opened tab, it pointed to the first search result.
We further analyze the error distribution between reasoning errors (the VLM/LRM mistranslates the instruction and refers to the wrong element) and grounding errors (the rewritten instruction is correct, but the grounding model still fails to click the right element). For the best model (o3 + GTA-1), 43.3\% of errors are grounding errors and 56.7\% are reasoning errors. This suggests that current VLMs and LRMs still lack the reasoning and planning abilities needed to robustly perform sequential instruction-following tasks.

\textbf{Does learning from real-world user data help?}
Fine-tuning GTA-1 marginally improved average progress and step accuracy but yielded no additional benefit when paired with VLMs and LRMs. These results show that the fine-tuned model better understands real user instructions, yet it still fails to generalize to instructions generated by VLMs and LRMs. The experiments suggest that finetuning grounding models on a small set of real user instructions provides minimal benefit, and collecting large-scale real user instructions remains a significant challenge.

\textbf{Limitations.}
RealWebAssist represents an important first step towards evaluating web agents on long-horizon, real-user tasks. However, it has several limitations. The first is participant scale and diversity. Collecting real-user data is expensive and time-consuming. The number of participants is comparable to prior works that use expert annotators \citep{lu2024weblinx}. However, we intend to increase user diversity in future versions of the benchmark. We will also open-source our data collection tools for community expansion of the dataset. Second, like prior benchmarks on real-world websites \citep{deng2024mind2web, cheng2024seeclick}, we constrain our evaluation to an offline setting to ensure reproducibility and safety. This is complementary to benchmarks that focus on interactive evaluation in sandbox environments (e.g., WebArena). We believe that web agents should be evaluated on both types of benchmarks to fully assess their capabilities. Lastly, the current setting does not allow dialogue between a user and the AI assistant, which we will explore in future work.

% RealWebAssist represents an important first step towards evaluating web agents on long-horizon, real-user tasks, but it has several limitations. The first is participant scale and diversity. Collecting real-user data is expensive and time-consuming. The number of participants is comparable to prior works that use expert annotators \citep{lu2024weblinx}. However, we intend to increase user diversity in future versions of the benchmark. We will also open-source our data collection tools for community expansion of the dataset. Second, like prior benchmarks on real-world websites \citep{deng2024mind2web, cheng2024seeclick}, we constrain our evaluation to an offline setting to ensure reproducibility and safety. In a long-horizon task, a single step deviation from the user's trajectory would lead to entirely different webpages and interactions. Thus, our offline evaluation assumes successful execution of prior steps. However, evaluating web agent planning methods, particularly model-based methods (e.g., \cite{putta2024agent}), ideally requires interactive simulation environments. RealWebAssist provides evaluation of sequential instruction following on real-world websites, which is complementary to benchmarks that focus on interactive evaluation in sandbox environments (e.g., WebArena). We believe that web agents should be evaluated on both types of benchmarks to fully assess their capabilities. Lastly, the current setting does not allow dialogue between a user and the AI assistant, which we will explore in future work.

\section{Conclusion}
In this paper, we present RealWebAssist, the first benchmark for evaluating web agents' ability to provide long-horizon web assistance with real-world users via sequential instruction-following. Our benchmark poses novel challenges, including spatial and temporal reasoning, planning, and adapting to user-specific routines. We conducted a comprehensive evaluation and analysis on multiple state-of-the-art GUI grounding models, VLMs, and LRMs, revealing critical limitations of them. We have also shown the limited benefit of finetuning models on real user data. Our benchmark, along with the well-annotated user instruction dataset, provides resources and diagnostic tools for further research on real-world web assistance. In future work, we plan to expand our human study to include more participants from various backgrounds, examine web assistance in interactive settings, and incorporate chat between users and web agents.

\section{Acknowledgements}
This work was supported by a research grant from Amazon. We thank Janice Chen for helpful discussions.

\bibliography{aaai2026}
\newpage
\begin{strip}
\centering
\vspace*{2.5em}
{\huge \textbf{Appendix}\par}
\vspace*{2.5em}
\end{strip}
\section{More experiment results}

\subsection{Full VLM \& LRM + Grounding results}

For the best three grounding models, GTA-1 \citep{yang2025gta1}, GUI-Actor \citep{wu2025gui} and UI-TARS \citep{qin2025ui}, we test their pairing with all the VLMs and LRMs. Table \ref{tab:categories} shows the full results. All the evaluation experiments are run on a single A100 GPU for 20 - 40 minutes. Finetuning GTA-1 model takes 4 hours on 4 A100 GPUs.

\begin{table*}[h!]
    \centering
    \begin{tabular}{p{2.5cm}|p{5cm}|>{\centering\arraybackslash}p{1.5cm}|
    >{\centering\arraybackslash}p{1.5cm}|
    >{\centering\arraybackslash}p{1.5cm}}
        \hline
        \multirow{3}{2.5cm}{VLM + GTA-1}
                & GPT-4o + GTA-1 & 8.4 & 23.5 & 72.7 \\
                & Qwen 2.5 72B + GTA-1 & 9.3 & 24.3 & 69.0 \\
                & Gemini 2.5 Flash + GTA-1 & 11.2 & 26.9 & 75.4 \\
        \hline
        \multirow{5}{2.5cm}{LRM + GTA-1}         
                  & Claude 3.7 Sonnet + GTA-1 & 12.1 & 26.7 & 68.8 \\
                  & Gemini 2.5 Pro + GTA-1 & 8.4 & 23.5 & 74.5 \\
                  & o1 + GTA-1 & 7.5 & 21.1 & 73.1 \\
                  & o3 + GTA-1 & 14.0 & 28.7 & 76.7 \\
                  & o4-mini + GTA-1 & 10.3 & 21.7 & 67.1 \\
        \hline
        \multirow{3}{2.5cm}{VLM + GUI-Actor}
                & GPT-4o + GUI-Actor & 6.5 & 18.0 & 67.0 \\
                & Qwen 2.5 72B + GUI-Actor & 9.3 & 21.4 & 64.9 \\
                & Gemini 2.5 Flash + GUI-Actor & 10.3 & 25.6 & 73.1 \\
        \hline
        \multirow{5}{2.5cm}{LRM + GUI-Actor}         
                  & Claude 3.7 Sonnet+ GUI-Actor & 7.5 & 18.5 & 63.9 \\
                  & Gemini 2.5 Pro + GUI-Actor & 9.3 & 24.0 & 73.2 \\
                  & o1 + GUI-Actor & 7.5 & 17.7 & 68.2 \\
                  & o3 + GUI-Actor & 12.1 & 27.4 & 74.0 \\
                  & o4-mini + GUI-Actor & 8.4 & 20.0 & 65.1 \\
        \hline
        \multirow{3}{2.5cm}{VLM + UI-TARS}
                & GPT-4o + UI-TARS & 6.5 & 20.8 & 67.3 \\
                & Qwen 2.5 72B + UI-TARS & 7.5 & 21.8 & 63.2 \\
                & Gemini 2.5 Flash + UI-TARS & 9.3 & 24.1 & 70.2 \\
        \hline
        \multirow{5}{2.5cm}{LRM + UI-TARS}         
                  & Claude 3.7 Sonnet + UI-TARS & 9.3 & 17.5 & 61.5 \\
                  & Gemini 2.5 Pro + UI-TARS & 7.5 & 23.4 & 71.6 \\
                  & o1 + UI-TARS & 6.5 & 18.5 & 66.0 \\
                  & o3 + UI-TARS & 12.1 & 27.2 & 72.4 \\
                  & o4-mini + UI-TARS & 7.5 & 19.4 & 62.5 \\
        \hline
    \end{tabular}
    \caption{Model Performance for pairing GTA-1, GUI-Actor and UI-TARS with all LRMs \& VLMs, including task success rate, average progress, and step accuracy. All results are in \%. }\label{tab:categories}
    \vspace{-10pt}
\end{table*}

\subsection{Experiment with different context lengths}

\begin{figure}[h]
  \centering \includegraphics[trim=0cm 15cm 32cm 0cm, clip, width=0.5\textwidth]{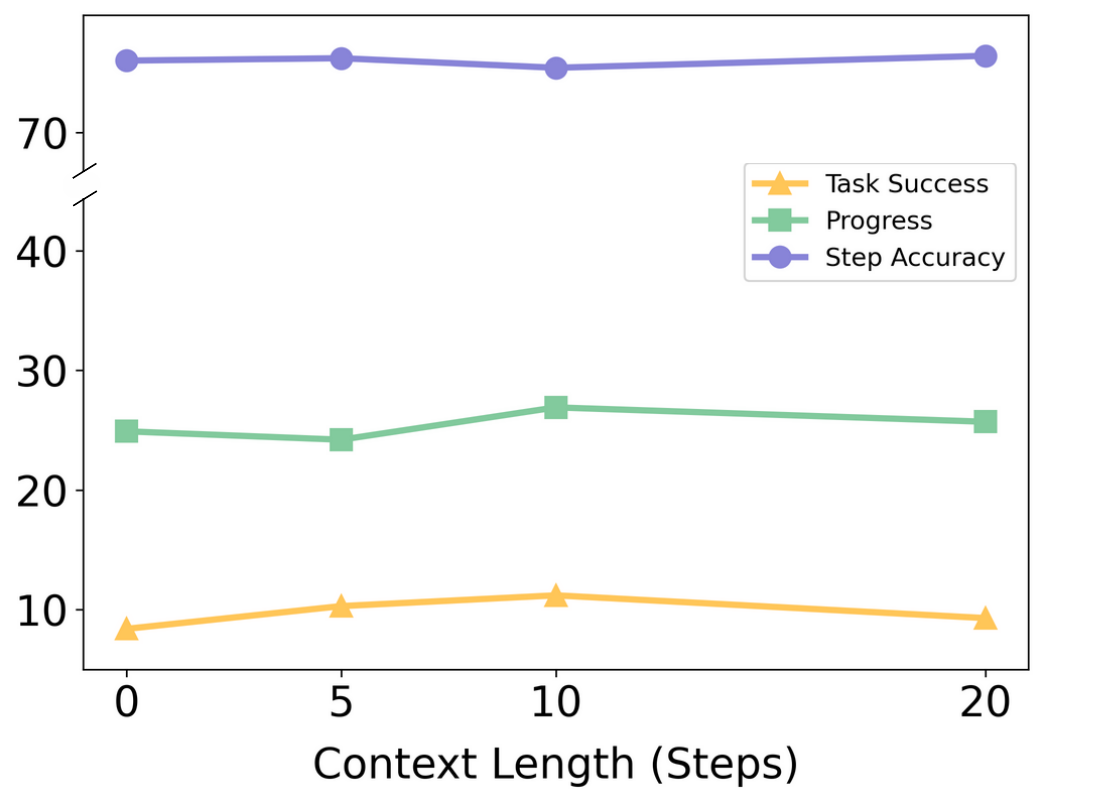}
  \caption{Effect of context length on Gemini 2.5 Flash + GTA-1. }
  \label{fig:context_length}
\end{figure}

We evaluated the best-performing VLM (Gemini 2.5 Flash) + GTA-1 with varying history context lengths, from no history to 20 steps. An ideal assistant should be able to leverage different kinds of historical context based on different instructions, ranging from no history to multi-task history context (e.g., for routine learning). As shown in Figure \ref{fig:context_length}, increasing context length also does not necessarily lead to better performance. Gemini 2.5 Flash + GTA-1 achieved the highest task success rate with a context length of 10, and increasing the context length further led to poorer performance. This suggest the limitation of VLM in effectively utilizing historical context for reasoning.

%We provide the results with the speech recognition results as input in Appendix~\ref{appendix:speech_to_text}.

\subsection{Effect of Speech Recognition Errors}
\label{appendix:speech_to_text}
All baseline experiments use the ground truth transcripts of user speech instructions as input to ensure that performance is not affected by errors in speech-to-text transcription. However, in real-world settings, instructions are often given via speech. To reflect this, we evaluated the effect of speech recognition on the agent's performance by using the transcripts generated from a state-of-the-art automatic speech recognition (ASR) model, Whisper Large-V3 \citep{radford2023robust}. Additionally, since users may not always be in quiet, controlled environments using a high-quality microphone like in our user experiment setup, we simulated noisy environments by injecting background noise with noise files from the Microsoft Scalable Noisy Speech
Dataset (MS-SNSD) dataset \citep{reddy2019scalable}, following \cite{ying2024siftomrobustspokeninstruction}. The noise files include people talking in the background and keyboard typing sounds. As shown in Table \ref{tab:audio_results}, using speech recognition resulted in a 1.9\% drop in task success rate, and having noisy speech resulted in a further 1.9\% drop. In contrast, the word error rate (WER) of the ASR results increased from 1.4\% (original speech) to 28.1\% (noisy speech), a much larger performance drop compared to the final task performance. This result suggests that reasoning the true meanings of speech instructions by leveraging context can help mitigate errors from ASR.

%the task performance could be improved by developing methods that better utilize context to mitigate speech recognition errors.
\begin{table}[htbp] 
    \centering 
    \begin{tabular}{p{3cm}|>{\centering\arraybackslash}p{1.2cm}|
                         >{\centering\arraybackslash}p{1.2cm}|
                         >{\centering\arraybackslash}p{1.2cm}}
    \hline
    Input Transcript & \makecell{Task\\Success} & Progress & \makecell{Step\\Accuracy} \\
    \hline
    Ground Truth & 10.3 & 21.7 & 66.4 \\
    Whisper Large-V3 & 8.4 & 20.9 & 65.5 \\
    Whisper Large-V3 (Noise) & 6.5 & 20.6 & 63.4 \\
    \hline
\end{tabular}
    \caption{Performance of GPT-4o + UGround-V1 using (1) ground-truth transcripts, (2) transcripts generated from original user speech by Whisper Large-V3, and (3) transcripts generated from noisy speech by Whisper Large-V3.}
    \label{tab:audio_results}
\end{table}

\section{Dataset Construction Details} \label{app:full}
\label{appendix:dataset_construction_details}
\textbf{Video Segmenting.} As shown in the video example \ref{appendix:video_example}, the interactive sessions are highly dynamic, and spoken instructions do not always align cleanly with specific screens or timesteps. Automatically segmenting instructions and matching them to corresponding webpages and actions using heuristics would risk significantly degrading data quality. Therefore, we manually segment the live sessions using video editing software to construct the final RealWebAssist dataset. All participants provided consent to have their speech recorded and included in this dataset.

\begin{figure*}[htbp]
    \centering
    \includegraphics[trim=0cm 18cm 15cm 0cm, clip, width=1.0\textwidth]{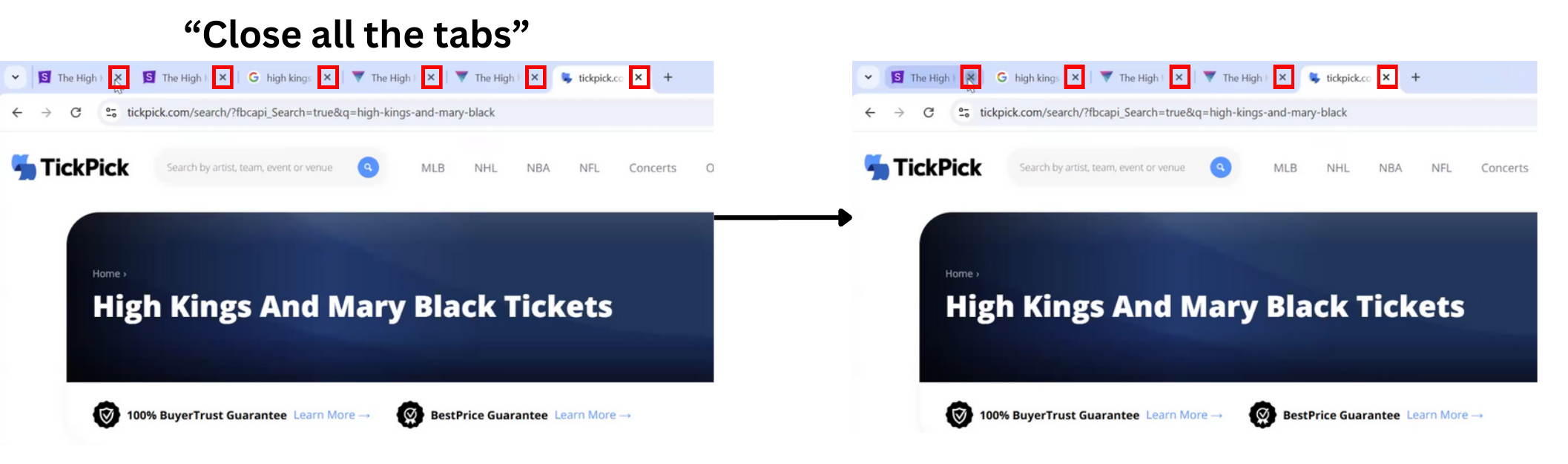}
    \vspace{-10pt}
    \caption{Example of annotated bounding boxes for an instruction. The red boxes represent the correct bounding boxes. The user gave the instruction ``Close all the tabs". For evaluation purposes, closing any of the tabs first is considered correct at each step, so all the x marks are labeled as correct at each step.}
    \label{fig:close_all_the_tabs}
    \vspace{-10pt}
\end{figure*}

\textbf{Bounding Box Labeling.} As shown in Figure \ref{fig:close_all_the_tabs}, certain instructions like ``close all the tabs" may correspond to multiple valid actions, since closing any of the tabs first would be reasonable. Therefore, we add bounding boxes to all of the elements that would be correct. The bounding boxes are drawn manually using a Python tool built with tkinter, and the clickable regions are determined by a visual inspection of the webpage. 

\section{More Dataset Details}

\subsection{Evaluation detail}

User instructions in RealWebAssist require different operations on the webpage, including clicking, scrolling and typing. We believe that action types other than clicking is trivial (for typing actions, the benchmark includes the step of finding the correct place to type instead of the actual typing process), so we only evaluate click‑type actions with annotated bounding boxes are scored; instructions like “scroll” remain in the history but are not counted in our metrics. Of the 1,885 instructions, 1,412 are scored, yielding 1,714 evaluated action steps (one screenshot per step). Tasks average 17.6 evaluated steps. 

\subsection{User behaviors}
\begin{figure*}[htbp]
    \centering
    \includegraphics[trim=11.3cm 2cm 11.3cm 2cm, clip, width=1.0\textwidth]{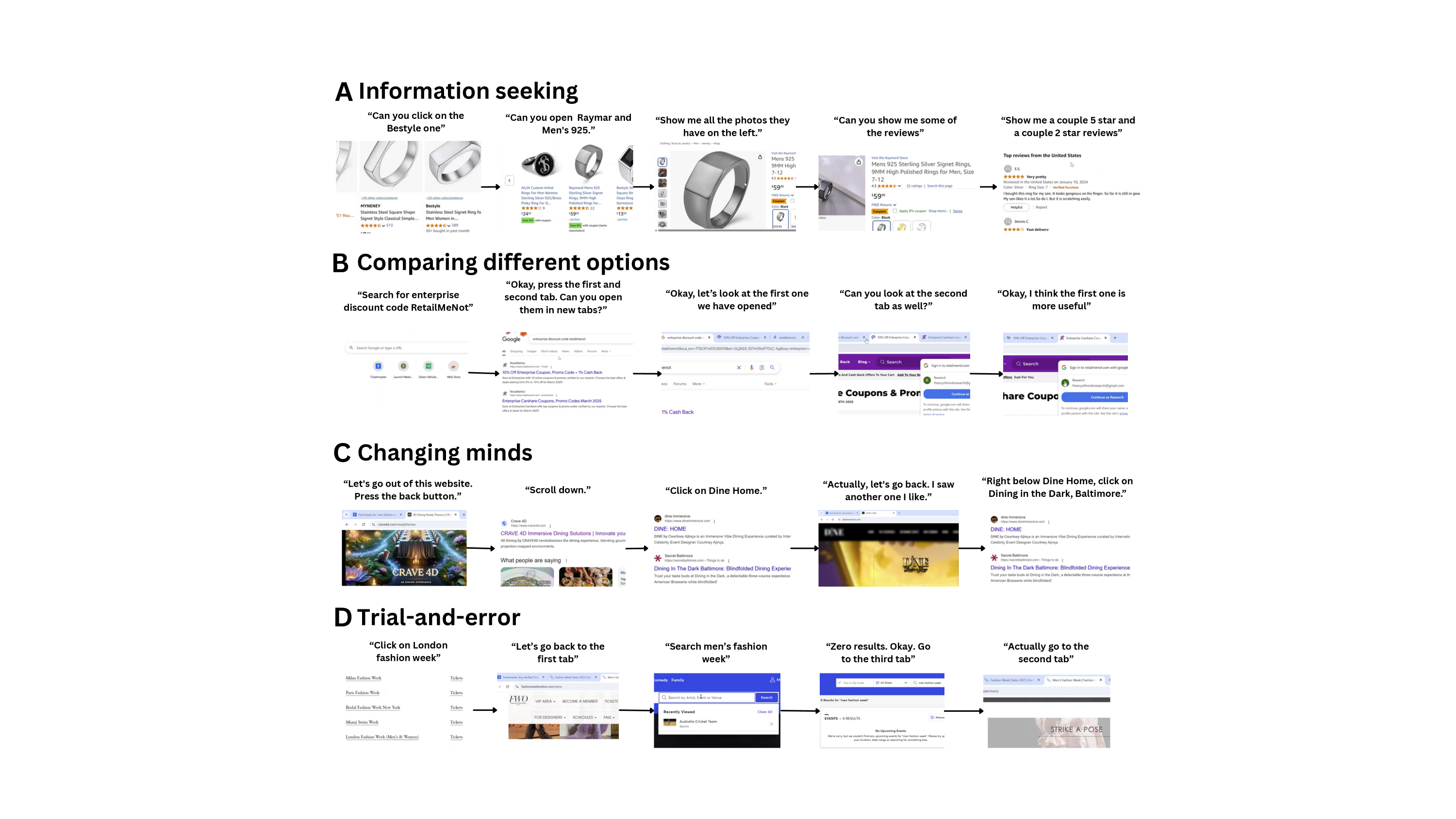}
    \vspace{-10pt}
    \caption{Example of rich user behaviors in RealWebAssist.}
    \label{fig:behaviors}
    \vspace{-10pt}
\end{figure*}

Figure \ref{fig:behaviors} shows diverse user behaviors in RealWebAssist not present in previous benchmarks. We include a zip file of the live recordings (including audio) from which the examples are taken.

\noindent \textbf{Information seeking} As Figure \ref{fig:behaviors}A shows, the user is seeking information from different aspects, like images and ratings, before they make the purchase decision.

\noindent \textbf{Comparing different options} Figure \ref{fig:behaviors}B shows the process of the user viewing two candidates and finally make the decision between them.

\noindent \textbf{Changing minds} In Figure \ref{fig:behaviors}C, the user is searching for some immersive dining experience. They are checking different restaurants and frequently change their minds when they see more options.

\noindent \textbf{Trial-and-error} As Figure \ref{fig:behaviors}D shows, the user has several unsuccessful attempts when searching for men's fashion week. They refer to previous searchs or initiate new ones to look for what they want.

\noindent These diverse behaviors increase the complexity of the web assistance: instead of clearly defined-goals, the user themselves are also actively collecting knowledge to make decisions, which requires web assistant to follow the user's mind and act accordingly.

\label{appendix:dataset_statistics}
\onecolumn
\subsection{Full List of Tasks}
\label{appendix:full_tasks}
\begin{longtable}{@{}r p{13cm}@{}}
\toprule
\textbf{Task \#} & \textbf{Description} \\
\midrule
\endfirsthead

\toprule
\textbf{Task \#} & \textbf{Description} \\
\midrule
\endhead

\bottomrule
\endfoot

\bottomrule
\endlastfoot

1 & Buy a gift for each of my three friends with a budget of \$100 \\
2 & Find and buy a birthday gift for a friend who loves tech, within a \$50 budget. \\
3 & Purchase a cute water bottle for everyday use, under \$15 \\
4 & Compare different laptops and buy one with the best review \\
5 & Purchase three home workout items under \$75 and compare their reviews before buying. \\
6 & Find and order a customized gift (e.g., engraved or personalized) for a friend’s graduation under \$60. \\
7 & Order a complete warm and durable winter outfit (jacket, gloves, and boots) under \$200. \\
8 & Get two sets of reusable grocery bags under \$20 total, checking for durability and eco-friendliness. \\
9 & Buy two wall paintings for a family house, one for a 13-year old boy, one for a 6-year old girl \\
10 & Purchase a set of colorful coffee mugs under \$20 with fun designs \\
11 & Buy a small easy-care indoor plant under \$15 and schedule delivery within three days \\
12 & Get a colorful umbrella for under \$30, making sure it’s big enough for two people \\
13 & Buy a set of scented candles under \$25, ensuring they have good reviews for long-lasting fragrance.\\
14 & Find and purchase a durable phone case under \$20 for an iPhone 14 Pro Max.\\
15 & Order a cozy throw blanket under \$30, checking for softness and warmth.\\
16 & Buy a set of three face masks (reusable \& breathable) under \$15.\\
17 & Get a wireless Bluetooth speaker under \$40 with good bass and waterproofing.\\
18 & Order a set of noise-canceling earplugs under \$15, ensuring they’re comfortable for sleep.\\
19 & Find and buy a compact travel pillow and eye mask set under \$30.\\
20 & Purchase a set of six kitchen towels under \$20 with high absorbency.\\
21 & Buy an adjustable desk lamp under \$35 with multiple brightness settings.\\
22 & Order a pack of 12 gel pens under \$15 in assorted colors with smooth writing.\\
23 & Purchase a waterproof picnic blanket under \$40, ensuring it’s easy to fold and carry.\\
24 & Buy a cute yet professional notebook under \$20 for journaling or work.\\
25 & Find and purchase a comfortable memory foam seat cushion under \$35 for long sitting hours.\\
26 & Order a set of reusable silicone food storage bags under \$25.\\
27 & Buy a pair of comfy indoor slippers under \$30 with high reviews for warmth and durability.\\
28 & Purchase a portable mini humidifier under \$40 with USB charging.\\
29 & Order a stylish travel makeup bag under \$25, ensuring it has multiple compartments.\\
30 & Find and order a surprise gift box for a friend who enjoys skincare, under \$50.\\
31 & Compare wireless earbuds and purchase the best-reviewed pair under \$100.\\
32 & Order a budget-friendly yet stylish smartwatch under \$75, ensuring good battery life.\\
33 & Find and order a high-quality mechanical keyboard under \$120, comparing typing feel and reviews\\
34 & Find and buy a useful desk gadget under \$40 for a friend who works from home\\
35 & Plan flights for a trip from US to Europe (at least two different countries) for 3 days, comparing different airlines to find the best deal.\\
36 & Plan a 5-day trip to Japan, booking both flights and hotels, taking into account customer reviews.\\
37 & Book a hotel for a weekend trip for a good price near the beach within the country, making sure you can cancel the trip at any time\\
38 & Plan a spontaneous weekend trip to a destination with cheap last-minute flights and good hotel deals, for hotel make sure it’s comfortable enough.\\
39 & Book a luxury hotel for a weekend at a city in the west US, pay attention to different services offered\\
40 & Plan a three-stop European trip in a single week, with flights and hotel for each place\\
41 & Book hotel for a family tour of four to a kid-friendly destination, with a hotel offering family amenities and breakfast included.\\
42 & Arrange a road trip across the US, booking rental cars and a mix of motels and boutique hotels along the route.\\
43 & Book a romantic beach getaway in Hawaii for two people, make sure it’s close to beach and have sea view\\
44 & Plan a family Disney Cruise, securing flights to Port Canaveral and a hotel near the theme parks before sailing.\\
45 & Arrange a wine country getaway, booking flights to Napa Valley, a rental car, and a vineyard hotel with wine-tasting experiences.\\
46 & Find flights and a convertible rental car for a coastal drive in Hawaii, staying in beachfront resorts along the way.\\
47 & Choose flights to a popular ski destination and secure a lodge or hotel under \$150/night.\\
48 & Book last-minute flights and a centrally located hotel in a major US city, focusing on deals under \$100/night with great city landscape view.\\
49 & Secure round-trip flights to a scenic South American city and book a comfortable hotel near local attractions.\\
50 & Pick flights from a major US airport to a warm city in Canada, with a hotel under \$100/night in the downtown area.\\
51 & Schedule flights and a boutique hotel stay in a city rich in history, aiming for under \$100/night in a central location.\\
52 & Arrange direct flights to a popular theme park region, booking a nearby hotel or hotel with easy transportation\\
53 & Schedule flights for a quick visit to a popular national park, booking a nearby lodge or hotel with scenic views.\\
54 & Book round-trip flights to a major Middle Eastern city and reserve a modern hotel near historic sites for under \$100/night\\
55 & Secure flights from the US to a tropical island, choosing a resort that offers water sports\\
56 & Find flights and a resort for a tropical vacation in Cancun, Mexico, focusing on all-inclusive options for relaxation\\
57 & Book flights to Cairo for a 5-day trip, then pick a hotel with a direct view of the Pyramids and free breakfast included\\
58 & Book a solo retreat to Kyoto, Japan, selecting a traditional ryokan stay with an onsen and authentic Japanese breakfast.\\
59 & Buy tickets for 2 people to an NBA Basketball game next weekend.\\
60 & Find and book tickets for a concert by a top artist in the nearest major city within the next three months.\\
61 & Search for a last-minute concert ticket and find the best available seat.\\
62 & Book 3 tickets for a rivalry match between two major sports teams\\
63 & Book 3 tickets for a unique or unusual event, such as a drag show, wrestling match, or haunted experience\\
64 & Purchase four tickets for a Broadway musical happening next month, aiming for orchestra seats if possible.\\
65 & Buy tickets for a family of 4 with 2 kids to a MLB game\\
66 & Find and book tickets to a popular stand-up comedy show in a western big city for the upcoming weekend, prioritizing seats near the front.\\
67 & Locate discounted tickets for a live theater performance in California this weekend\\
68 & Search for an NFL game next month and buy two tickets in a mid-priced seating section for some eastern teams\\
69 & Identify and reserve tickets for a children’s matinee performance at a local venue, comparing any available family packages or group discounts.\\
70 & Secure seats for a must-see hockey match, comparing “Best Seat” options.\\
71 & Find tickets for a classical music or orchestra concert in the nearest major city next month, aiming for seats with a good view of the stage.\\
72 & Buy tickets for two people to an English Premier League soccer match in London city center next weekend.\\
73 & Find and purchase tickets to a major electronic music festival in Las Vegas within the next two months.\\
74 & Book seats for a stand-up comedy show in downtown Chicago next month, make sure the location is in city center.\\
75 & Search for tickets to a top-tier cricket match in Sydney next month, aiming for seats that offer a good view of the pitch\\
76 & Locate a family-friendly musical performance near your city for next month.\\
77 & Purchase two tickets to an upcoming rugby match in Dublin next month, making sure seats are in a central section and remain under.\\
78 & Find a highly rated ballet or opera production in Paris within the next two months, choose the seat in the second floor if available\\
79 & Find tickets to a major fashion event, such as a runway show or fashion week experience.\\
80 & Look for tickets to a themed immersive dining experience (e.g., murder mystery dinner, fantasy-inspired restaurant)\\
81 & Book tickets for UEFA soccer game between two Spanish teams for the next week\\
82 & Book a ticket for a rooftop movie screening or outdoor film festival in a major city.\\
83 & Find tickets for an esports event and compare standard vs. premium seating options.\\
84 & Book a ticket for a ``silent disco" event in a city of your choice.\\
85 & secure two tickets to a major MLB game in a well-known ballpark anywhere in the U.S. next month, opting for seats along the first baseline.\\
86 & Find and book tickets for a large-scale country music festival occurring in the southern U.S. within the next two months, focusing on general admission passes.\\
87 & Purchase seats for a top-tier college football rivalry game taking place within the next six weeks, ensuring you can view the marching band’s performance easily.\\
88 & Reserve tickets to a major NHL match in the next two months, choosing seats close to the ice.\\
89 & Book passes for a nationally touring art exhibition or immersive art experience within the next two months, ensuring weekend availability.\\
90 & Secure seats for a top-rated Broadway musical in New York City, making sure the date aligns with a Saturday evening performance.\\
91 & Reserve a spot for a special museum or cultural center night event (e.g., “Night at the Museum” or themed after-hours) in a major U.S. city within the next two months.\\
92 & Find the best deal on a new smartphone (latest model iPhone or Samsung)\\
93 & Find the best dinner deal for two using food delivery apps\\
94 & Purchase an outfit for a formal event within a \$150 budget\\
95 & Buy a high-quality gaming chair for under \$250\\
96 & Find and book the best available concert tickets for a top artist in your city\\
97 & Book tickets for a live theater performance and find a pre-show dinner reservation\\
98 & Plan a sports game outing for two within a \$150 budget\\
99 & Plan a weekend getaway for two within a \$500 budget\\
100 & Organize a one-day itinerary for a solo traveler in a major city\\
101 & Compare car rental options for a 5-day road trip\\
102 & Find and book a local escape room challenge for a group of four\\
103 & Plan a movie night with discounted tickets and snacks\\
104 & Find a highly-rated sushi restaurant and order a meal for delivery\\
105 & Plan a surprise birthday dinner at a fine dining restaurant\\
106 & Order a late-night snack under \$15 for delivery\\
107 & Book a luxury hotel staycation for a weekend\\
\end{longtable}

\subsection{Full List of Websites}\label{app:websites}
\begin{footnotesize}
\begin{longtable}{llll}
\toprule
\textbf{Name} & \textbf{URL} & \textbf{Task Type} \\
\midrule
\endfirsthead
\toprule
\textbf{Name} & \textbf{URL} & \textbf{Task Type} \\
\midrule
\endhead
\bottomrule
\endfoot
ACL Festival & aclfestival.com & Entertainment \\
Amazon & amazon.com & Shopping \\
Ammoora & ammoora.com & Entertainment \\
Apple & apple.com & Shopping \\
Artechouse & artechouse.com & Entertainment \\
Atom Tickets & atomtickets.com & Entertainment \\
Best Buy & bestbuy.com & Shopping \\
Adidas Arena & billetterie.adidasarena.com & Entertainment \\
Broadway & broadway.com & Entertainment \\
Charm City Clue Room & charmcityclueroom.com & Entertainment \\
City Pass & citypass.com & Travel Planning \\
CN Tower & cntower.ca & Travel Planning \\
Colorado Tourism & colorado.com & Travel Planning \\
Corsair & corsair.com & Shopping \\
Coupon Follow & couponfollow.com & Shopping \\
Crave 4D & crave4d.com & Entertainment \\
Dine Immersive & dineimmersive.com & Food \\
Disney Cruise & disneycruise.disney.go.com & Travel Planning \\
DoorDash & doordash.com & Food \\
Drone and DSLR & droneandslr.com & Shopping \\
Enterprise & enterprise.com & Travel Planning \\
ESCharts & escharts.com & Entertainment \\
ETIX & etix.com & Entertainment \\
Eventbrite & eventbrite.com & Entertainment \\
Expedia & expedia.com & Travel Planning \\
Fashion Week Online & fashionweekonline.com & Entertainment \\
Fever Up & feverup.com & Entertainment \\
Google & google.com & Travel Planning \\
Google Maps & google.com/maps & Travel Planning \\
Live Nation & livenation.com & Entertainment \\
Library of Congress & loc.gov & Travel Planning \\
LoL Esports & lolesports.com & Entertainment \\
MLB & mlb.com & Entertainment \\
MLB Tickets & mlb.tickets.com & Entertainment \\
NYICFF & nyicff.org & Entertainment \\
OpenTable & opentable.com & Food \\
Postmates & postmates.com & Food \\
Rakuten & rakuten.com & Shopping \\
Reddit & reddit.com & Entertainment \\
Retail Me Not & retailmenot.com & Shopping \\
Road Trip USA & roadtripusa.com & Travel Planning \\
Samsung & samsung.com & Shopping \\
San Lorenzo DC & sanlorenzodc.com & Food \\
Screen Daily & screendaily.com & Entertainment \\
Secret Baltimore & secretbaltimore.com & Travel Planning \\
Secret Lab & secretlab.co & Shopping \\
Smithsonian Sleepovers & smithsoniansleepovers.org & Entertainment \\
StubHub & stubhub.com & Entertainment \\
The Bureau Fashion Week & thebureaufashionweek.com & Entertainment \\
The Meltdown & themeltdown.com & Entertainment \\
The UFL & theufl.com & Entertainment \\
Ticketmaster & ticketmaster.com & Entertainment \\
Ticketmaster France & ticketmaster.fr & Entertainment \\
Ticket Web & ticketweb.com & Entertainment \\
TickPick & tickpick.com & Entertainment \\
TripAdvisor & tripadvisor.com & Travel Planning \\
Two Step Inn & twostepinn.com & Entertainment \\
Two Step Inn Frontgate & twostepinn.frontgatetickets.com & Entertainment \\
Uber & uber.com & Travel Planning \\
Uber Eats & ubereats.com & Food \\
Viator & viator.com & Travel Planning \\
Vivid Seats & vividseats.com & Entertainment \\
Washington Tourism & washington.org & Travel Planning \\
Yelp & yelp.com & Food \\
Zara & zara.com & Shopping \\
\bottomrule
\end{longtable}
% \caption{Website List with Task Categories}
\end{footnotesize}
\twocolumn
\subsection{Word Frequency}

\begin{figure}[htbp]
    \centering
    \includegraphics[trim=0cm 0cm 0cm 0cm, clip, width=0.5\textwidth]{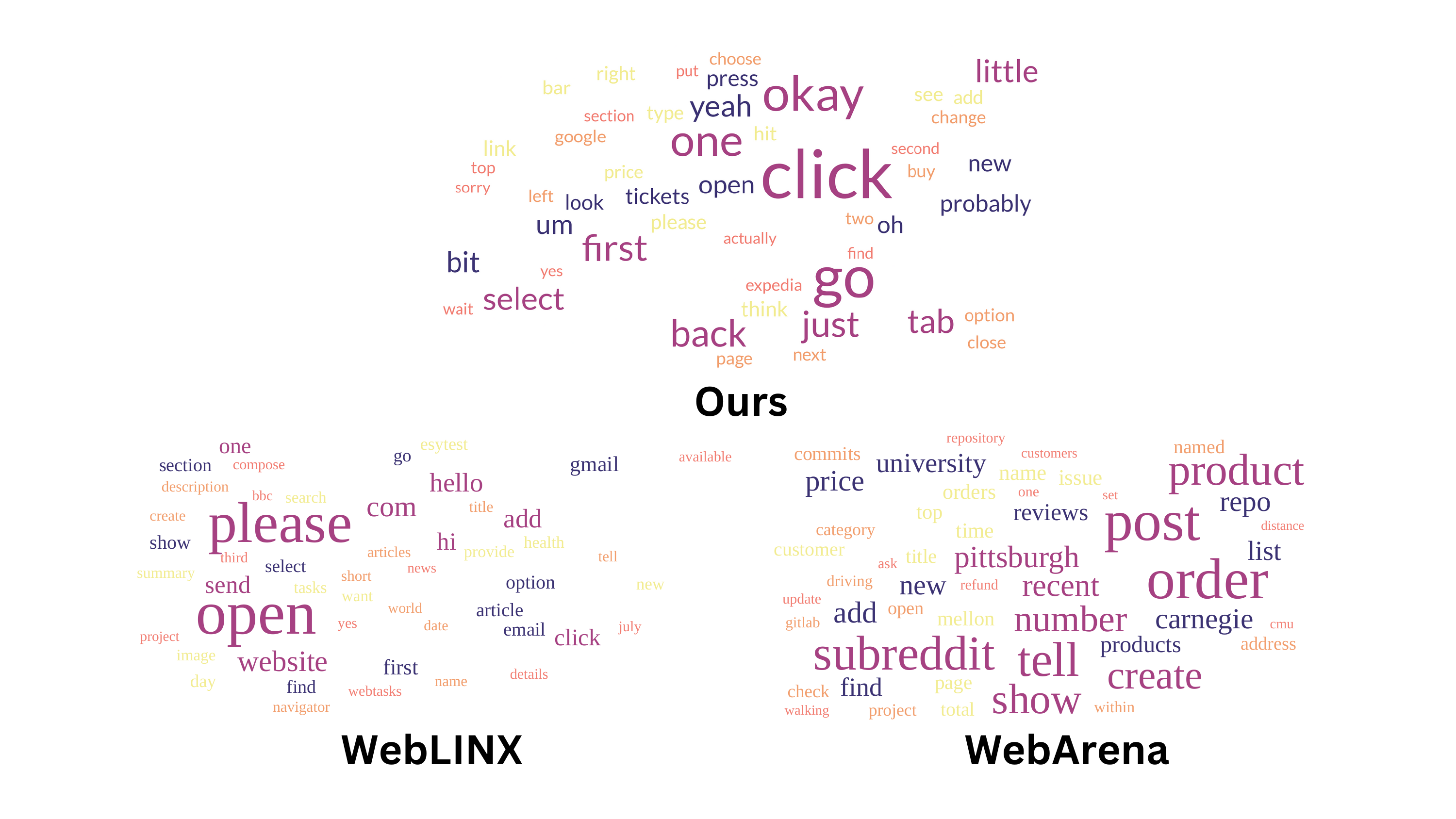}
    \caption{Word Cloud of the most frequent words in RealWebAssist v.s. common benchmarks WebLINX and WebArena.}
    \label{fig:word_cloud}
\end{figure}

Figure \ref{fig:word_cloud} compares the most frequent instruction words in RealWebAssist with those from two common benchmarks, WebLINX and WebArena. The vocabulary used in RealWebAssist is more informal, as the dataset comes from natural spoken instructions. The tone is also more informal and conversational compared to WebLINX and WebArena. 

\section{Instructions for the participants}
\begin{tcolorbox}
Thank you for participating in our study! You’ll be guiding another person who is controlling the computer on your behalf. Imagine you are \textbf{helping a friend navigate a website remotely}, giving step-by-step instructions to complete a task. Feel free to interpret the task as you see fit.
Here are some guidelines to keep in mind:
\begin{itemize}
  \item Give instructions \textbf{as naturally as possible}, just like you would in real life.
  \item You don’t have to be overly precise—\textbf{say what feels natural}.
  \item You can only give one instruction at a time. After the operator follows your instruction, \textbf{wait for them to complete it} before giving the next step.

  \item Keep your instructions \textbf{clear and concise}, but don’t stress too much about exact wording—just say what comes to mind!
  \item You are \textbf{allowed} to instruct the operator to use Google to search for things.
\end{itemize}
\end{tcolorbox}

\section{Video Example}
\label{appendix:video_example}
\begin{tcolorbox}[colback=gray!10, colframe=gray!80, title=A sample raw recording can be viewed via the link below (audio included)]
\url{https://youtu.be/CcyIt9tr5qo}
\end{tcolorbox}

\end{document}